%% file: main.tex
\documentclass[sigconf]{acmart}
\acmConference[DAC '26]{}{July 26--29, 2026}{Long Beach, CA}

\copyrightyear{2026}
\acmYear{2026}
\setcopyright{cc}
\setcctype{by}
\acmConference[DAC '26]{63rd ACM/IEEE Design Automation Conference}{July 26--29, 2026}{Long Beach, CA, USA}
\acmBooktitle{63rd ACM/IEEE Design Automation Conference (DAC '26), July 26--29, 2026, Long Beach, CA, USA}
\acmDOI{10.1145/3770743.3804365}
\acmISBN{979-8-4007-2254-7/2026/07}

\usepackage{enumitem}
\setlist{nosep, topsep=2pt, partopsep=0pt, parsep=0pt, itemsep=1pt}

\usepackage{amssymb}
\usepackage{amsmath}
\usepackage{booktabs}
\usepackage{stfloats}
\usepackage[linesnumbered,ruled, vlined]{algorithm2e}
\usepackage{tikz}
\usetikzlibrary{decorations.pathreplacing}
\usepackage{arydshln}
\usepackage{enumitem}
\usepackage{multirow}
\usepackage{epsfig}
\usepackage{multicol,lipsum,xparse}
\usepackage{caption}
\usepackage{subcaption}
\usepackage{stmaryrd}
\usepackage{hyperref}
\usepackage{tikz}
\usepackage{pifont}
\usepackage{cleveref}
\usepackage{xspace}
\usepackage{amsmath,amssymb,amsfonts}
\usepackage{graphicx}
\usepackage{textcomp}
\usetikzlibrary{positioning, shapes.geometric, arrows.meta}
\usepackage[ruled,linesnumbered]{algorithm2e}
\usepackage{adjustbox}
\usepackage{xcolor}
\usepackage[dvipsnames]{xcolor}
\usepackage{siunitx}

\usepackage{svg}
\begin{document}
 
\newcommand{\cunxi}[1]{\textcolor{red}{[\small cunxi: ~#1]}}

\newcommand{\zhan}[1]{\textcolor{blue}{[\small zhan: ~#1]}}

\newcommand{\cchen}[1]{\textcolor{green}{[\small chen: ~#1]}}

\newcommand{\jiaqi}[1]{\textcolor{brown}{[\small jiaqi: ~#1]}}

\newcommand{\yutung}[1]{\textcolor{MidnightBlue}{[\small yutung: ~#1]}}

\newcommand{\guoheng}[1]{\textcolor{purple}{[\small guoheng: ~#1]}}


\title{TOPCELL: Topology Optimization of Standard Cell via LLMs}
\newcommand{\name}{\textsf{TOPCELL}\xspace}
\newcommand{\fixme}[1]{\textcolor{red}{\small [~#1~]}}
\author{Zhan Song}
\affiliation{%
  \institution{University of Maryland}
  \city{College Park}
  \country{USA}}
\email{zhansong@umd.edu}

\author{Yu-Tung Liu}
\affiliation{%
  \institution{University of Maryland}
  \city{College Park}
  \country{USA}}
\email{liuyt@umd.edu}

\author{Chen Chen}
\affiliation{%
  \institution{University of Maryland}
  \city{College Park}
  \country{USA}}
\email{chenchen099707@gmail.com}

\author{Guoheng Sun}
\affiliation{%
  \institution{University of Maryland}
  \city{College Park}
  \country{USA}}
\email{ghsun@umd.edu}

\author{Jiaqi Yin}
\affiliation{%
  \institution{University of Maryland}
  \city{College Park}
  \country{USA}}
\email{jyin629@umd.edu}

\author{Chia-tung Ho}
\affiliation{%
  \institution{NVIDIA}
  \city{Santa Clara}
  \state{CA}
  \country{USA}}
\email{markf12764@gmail.com}

\author{Ang Li}
\affiliation{%
  \institution{University of Maryland}
  \city{College Park}
  \country{USA}}
\email{angliece@umd.edu}

\author{Haoxing Ren}
\affiliation{%
  \institution{NVIDIA}
  \city{Austin}
  \state{TX}
  \country{USA}}
\email{markren@agentrys.ai}

\author{Cunxi Yu}
\affiliation{%
  \institution{University of Maryland}
  \city{College Park}
  \country{USA}}
\email{cunxiyu@umd.edu}

\renewcommand{\shortauthors}{Song, et al.}


\newcommand*\circled[1]{\raisebox{.4pt}
                    {\tikz[baseline=(char.base)]{
            \node[shape=circle,draw,inner sep=1pt, style={fill=white, text=black}, scale=0.75] (char) {\textbf{#1}};}}}
\newcommand{\CY}[1]{\textcolor{red}{\textbf{CY:} #1}}

\begin{abstract}
Transistor topology optimization is a critical step in standard cell design, directly dictating diffusion sharing efficiency and downstream routability. However, identifying optimal topologies remains a persistent bottleneck, as conventional exhaustive search methods become computationally intractable with increasing circuit complexity in advanced nodes. 
This paper introduces \name, a novel and scalable framework that reformulates high-dimensional topology exploration as a generative task using Large Language Models (LLMs). We employ Group Relative Policy Optimization (GRPO) to fine-tune the model, aligning its topology optimization strategy with logical (circuit) and spatial (layout) constraints. 
Experimental results within an industrial flow targeting an advanced 2nm technology node demonstrate that \name significantly outperforms foundation models in discovering routable, physically-aware topologies. When integrated into a state-of-the-art (SOTA) automation flow for a 7nm library generation task, \name exhibits robust zero-shot generalization and matches the layout quality of exhaustive solvers while achieving an \textbf{85.91$\times$} speedup.
\end{abstract}

\maketitle

\input{01sec-intro}
\input{02sec-motivation}
\input{03sec-background}
\input{04sec-approach}
\input{05sec-results}
\input{06-sec-conclusion}
\pagebreak

\small
\bibliographystyle{ACM-Reference-Format}
\bibliography{ref.bib}

\end{document}

%% file: 01sec-intro.tex
\section{Introduction}\label{sec:intro}

Standard cells are the fundamental building blocks of modern Application-Specific Integrated Circuit (ASIC) designs. Their Power, Performance, and Area (PPA) metrics directly impact overall system-level efficiency. At advanced technology nodes, the manual effort required to design and optimize these cells is substantial, creating a significant bottleneck in the design cycle \cite{cheng2025standard,liu2024maptune,liu2026maptune}. This establishes a critical need for robust automation solutions. 
As transistor counts per cell increase in advanced technologies, the demand for scalable standard cell design automation becomes more urgent \cite{guo2025multi}.


Topology optimization is a critical stage in standard cell design, as it directly influences the layout compactness, routing feasibility, and parasitic characteristics. The primary objective is to find a transistor arrangement that maximizes diffusion sharing where adjacent transistors share a single source/drain contact. This diffusion sharing is a fundamental technique for achieving an efficient layout \cite{, chung2024optimal}.
Recent SOTA framework\cite{chengso3} approaches this problem by transforming circuit netlists into hierarchical series-parallel trees and recursively exploring topology permutations. However, the complexity inherent in this recursive exploration means that as transistor counts increase, the approach quickly becomes intractable.


Recent breakthroughs in LLMs have demonstrated impressive capabilities for logical reasoning and processing complex data \cite{achiam2023gpt, guo2025deepseek,yu2025autonomous}. This has spurred new applications within Electronic Design Automation (EDA), including hardware code generation \cite{liu2023verilogeval, liu2024rtlcoder, chen2025menter}, hardware program repair \cite{xu2024automated, yao2024location}, design space exploration \cite{fu2023gpt4aigchip, wang2024chatcpu}, and EDA tool evolution \cite{yu2025autonomous}.
However, the application of LLMs to topology optimization remains largely unexplored. This process is traditionally characterized by extensive manual effort, and current automated approaches often struggle with scalability. 
We posit that the ability of LLMs to rapidly reason over complex inputs like netlists holds significant promise for overcoming these limitations.

This work investigates a crucial question: \textit{can an LLM refine, or even discover, physically aware transistor topologies for standard cells without resorting to an exhaustive search?} Given the exponential complexity of conventional methods, we present \name, an LLM-driven framework for standard-cell topology optimization. \name introduces a scalable approach by formulating topology synthesis as an end-to-end policy optimization problem. Taking a standard-cell netlist as input, \name utilizes an LLM policy to autonomously propose physically-aware topology modifications. This process allows it to efficiently explore the design space, optimizing for design–technology co-optimization (DTCO) objectives under advanced technology nodes and discovering high-quality topologies.
To summarize, our key contributions are as follows:
\begin{enumerate}
    \item \textbf{A Novel LLM-Driven Framework for Topology Optimization:} We introduce \name, a framework that leverages LLMs for standard cell topology optimization in advanced technology. The LLM policy is trained using a reward function derived from placement and routing (P\&R) feedback.

    \item \textbf{Autonomous Topology Discovery via GRPO:} By pioneering the application of GRPO for standard cell optimization, \name achieves superior efficiency. This enables automated discovery of high-quality, physically-aware topologies.

    \item \textbf{Comprehensive Evaluation of Performance and Generalization:} Extensive experiments demonstrate that \name drastically outperforms much larger, general-purpose foundation models. The framework also exhibits strong zero-shot generalization, scaling from its 3-input, 2nm training data to generate high-quality topologies for more complex standard cells in 7nm technology nodes.

    \item \textbf{Massive Acceleration of SOTA Automation Frameworks:} When integrated into an open-source, SOTA standard cell design automation tool, \name produces layouts of comparable quality while achieving an average speedup of \textbf{85.91x}.
\end{enumerate}

%% file: 02sec-motivation.tex
\begin{figure}[!ht]
    \centering
    \includegraphics[width=0.85\linewidth]{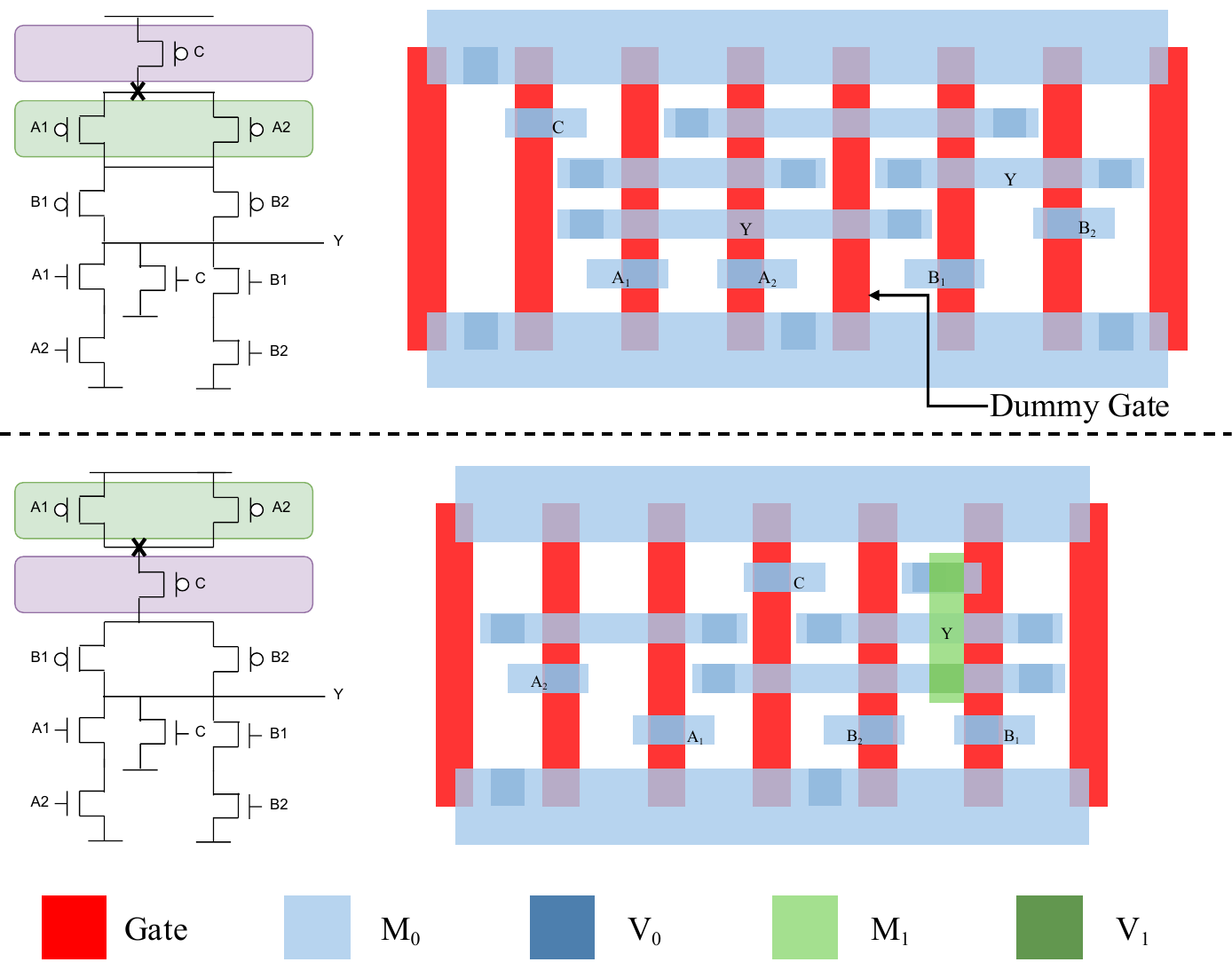}
    \caption{Comparison of 7$nm$ AOI221\_X1 layouts before (Top) and after (Bottom) topology optimization by \name. The optimized topology doesn't require a dummy gate insertion, leading to a smaller area. 
    This demonstrates the effectiveness of \name for improving cell layout quality.
    }
    \label{fig:layout}
\end{figure}

\section{Motivating Case Study}\label{sec:motivation}

The transistor topology is a critical factor that impacts the routability and layout quality of standard cells directly. For example, consider the AOI221\_X1 gate shown in Fig.~\ref{fig:layout}, which implements the Boolean function $Y = \overline{(A1 \cdot A2) + (B1 \cdot B2) + C}$. The suboptimal topology for this gate, shown at the top, can lead to a poor layout, creating diffusion gaps that necessitate inserting a dummy gate and wasting valuable cell area. \name avoids this by restructuring the transistor networks, as shown in the highlighted swap region. This reconfiguration yields the alternative, functionally-equivalent topology at the bottom. The new topology enables more effective diffusion sharing, which directly reduces cell area and improves pin accessibility, thereby enhancing downstream routability.

Discovering these beneficial topologies, however, is a major design bottleneck in standard cell design automation. Prior SOTA work, such as \textit{SO3-Cell}~\cite{chengso3}, relies on recursive or exhaustive design-space exploration, which suffers from exponential complexity as transistor count increases. For example, \textit{SO3-Cell} requires nearly 10 hours to find an optimal topology for AOI222\_X1\_SH which contains 12 transistors, while \name discovers the optimal topology in just 2 seconds. This massive computational burden renders search-based methods fundamentally impractical for larger libraries or complex, higher-input Boolean functions in advanced nodes.

%% file: 03sec-background.tex
\section{Background}\label{sec:background}

\subsection{LLM Applications in Standard Cell Design}

LLMs are increasingly being explored to automate labor-intensive aspects of standard cell design. For instance, LogicCraft~\cite{lin2025logiccraft} translates Boolean truth tables into SPICE netlists of standard cells using an LLM for combinational logic optimization. In the physical design domain, LLMs have been used as autonomous agents to refine layout parameters and optimize cell-level PPA metrics~\cite{ho2024large}. While these applications show great promise, they focus on either high-level logic implementation or post-topology layout refinement. Consequently, the fundamental potential for LLMs to systematically explore and optimize novel standard cell topologies, a task that defines the core performance envelope, remains largely unexplored.


\subsection{LLM Alignment via Reinforcement Learning}

Reinforcement Learning (RL) has emerged as a key technique for aligning LLMs with complex objectives. This process refines the model, steering it from a simple next-token predictor toward optimizing a goal-oriented policy \cite{guo2025deepseek}. 
Such alignment is essential for tasks with intricate constraints where Supervised Fine-Tuning (SFT) alone falls short. The core limitation of SFT is its reliance on a high-quality dataset pre-labeled optimal examples, which are often difficult or impossible to procure. 
RL overcomes this by allowing the model to learn an effective policy through iterative exploration, guided by feedback from a specialized reward model\cite{rajani2025scalpel}.


GRPO \cite{shao2024deepseekmath} is a stable, on-policy RL algorithm for LLM training that retains the vital exploratory capabilities of Proximal Policy Optimization (PPO)\cite{schulman2017proximal} while adopting a simpler, more memory-efficient design that eliminates the need for an explicit, unstable value function. This synthesis of on-policy exploration, stability, and efficiency makes GRPO a compelling choice for complex EDA tasks, such as RTL generation and buffer insertion \cite{wang2025verireason, hsiao2025buffalo}.

%% file: 04sec-approach.tex
 
\section{Approach}\label{sec:approach}

\subsection{Overview}
\begin{figure*}[!htb]
    \centering
    \includegraphics[width=0.95\linewidth]{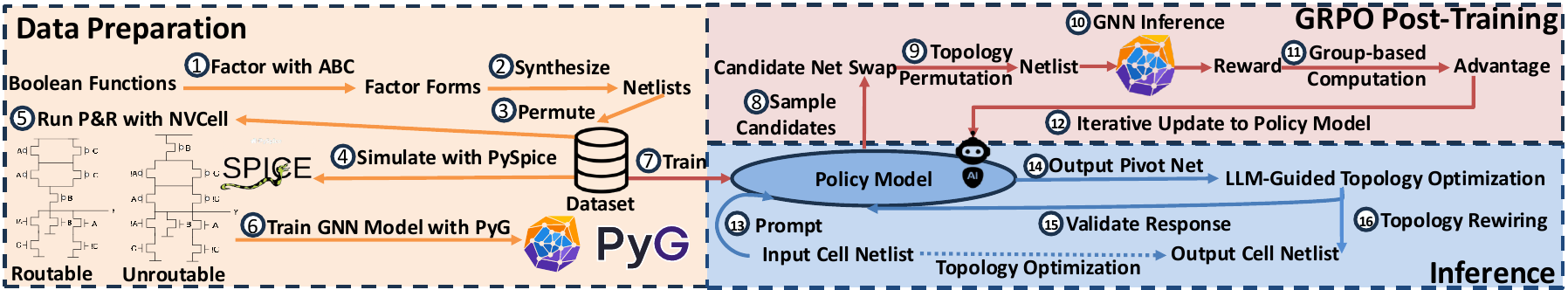}
    \caption{The \name workflow, showing Data Preparation (Left) to build a topology-diverse dataset and a GNN model for routability reward, and the main process (Right), which includes GRPO Post-Training (Top) to update the LLM policy and Inference (Bottom) to optimize a given cell topology.}
    \label{fig:workflow}
\end{figure*}

    \name introduces a novel, LLM-driven workflow to optimize standard cell topology, as depicted in Fig.~\ref{fig:workflow}. The main process consists of two stages, detailed on the right side of the figure. The \textbf{post-training stage} (Top) focuses on the policy update process. In this stage, given standard cell topologies in the training set (see Section~\ref{subsec:data}), the LLM policy suggests multiple candidate permutations. A Graph Neural Network (GNN) model, trained using \textit{NVCell 2} P\&R feedback\cite{ren2021nvcell, ho2023nvcell}, evaluates the routability of these candidates (see Section~\ref{subsec:gnn}). Finally, the GRPO algorithm leverages group-based advantages to iteratively update the LLM policy (see Section~\ref{subsec:grpo}). The \textbf{inference stage} (Bottom) utilizes the trained LLM policy. The policy proposes a physically-aware pivot net swap given an input standard cell netlist, which is then validated and executed by the LLM-Guided Topology Permutation module (see Section~\ref{subsec:swap}).


\subsection{LLM-Guided Topology Permutation}\label{subsec:swap}
\input{algorithm/dns}
\begin{figure}[!t]
    \centering
    \includegraphics[width=0.85\linewidth]{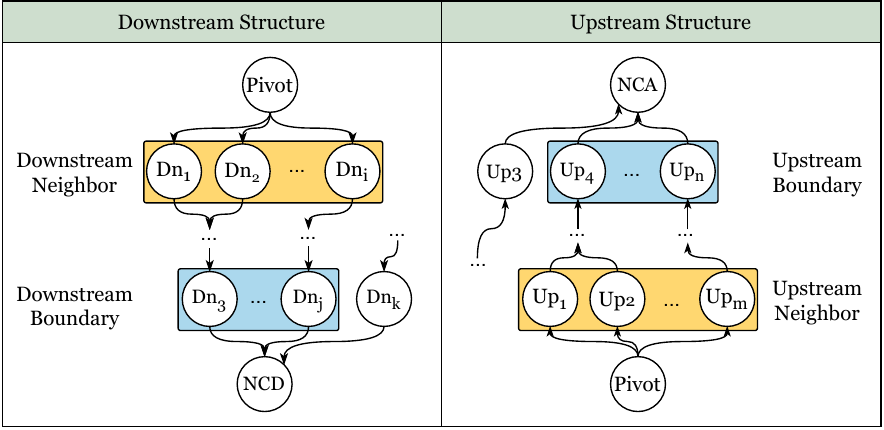}
    \caption{Structure of the Swap Region in LLM-Guided Topology Permutation. The region is bounded by NCA and NCD.}
    \label{fig:structure}
\end{figure}

\name performs topology modifications using a deterministic LLM-Guided Topology Permutation algorithm. We prompt an LLM to identify a single \textbf{pivot net} $pivot$ where a local swap is likely to improve routability (steps \circled{13}, \circled{14} in Fig.~\ref{fig:workflow}). The key to this approach is its decoupling of high-level strategic selection (the LLM) from low-level, graph-based execution (the algorithm). This separation significantly reduces the reasoning burden on the LLM and guarantees the \textbf{functional equivalence} of the resulting netlist.

As detailed in Alg.~\ref{alg:net_swapping}, the procedure unfolds in three main phases. First, the algorithm validates the LLM-selected $pivot$ (Step \circled{15}); if the net is invalid for swapping, the policy model is prompted to generate a valid alternative. Second, it defines a bounded \textbf{swap region} $\Delta$ by identifying the \textbf{Nearest Common Ancestor (NCA)} and \textbf{Nearest Common Descendant (NCD)} of the pivot net's neighbors (Fig.~\ref{fig:structure}). Third, the algorithm rewires the source-drain connections within $\Delta$ (step \circled{16}). As illustrated in Fig.~\ref{fig:layout} (Left), this process transforms the suboptimal topology within the colored swap region $\Delta$ (Top), centered on the $pivot$ (marked by a cross), into an equivalent one with optimized standard cell layout (Bottom).

\subsection{Data Preparation}\label{subsec:data}
We construct a dataset that exhaustively covers all three-input, single-output Boolean functions. Each standard cell topology is processed through a full P\&R flow, and unroutable topologies are used to train and evaluate LLMs. We intentionally restrict this dataset to 3 inputs, as the 4-input function space comprises $2^{2^{4}} = 65536$ possibilities. Exhaustively processing this much larger space through an end-to-end P\&R flow is computationally prohibitive.

\subsubsection{Boolean Factoring}

We begin by enumerating all $2^{2^{3}} = 256$ possible three-input Boolean functions. Excluding the two constant functions ($0$ and $1$), this leaves $254$ non-trivial functions. For each function, we first derive its canonical sum-of-products (SOP) representation via minterm expansion from its truth table. This SOP representation serves as the unoptimized, initial base.

Next, we compute an optimized multi-level factored form for each function using \textit{ABC}~\cite{brayton2010abc} (step \circled{1}). This step is critical because Boolean factoring exposes serial-parallel logic structures that map naturally to efficient CMOS implementations. And the resulting reduction in literal count generally correlates with a lower transistor count, leading to improved PPA metrics for the final cell~\cite{xiao2023minitntk, calvino2023improving}.


\subsubsection{Topology Enumeration}
Our process begins with a seed SPICE netlist synthesized from each optimized factored form (step \circled{2}). To explore the diverse topologies for a given Boolean function, we then use our LLM-Guided Topology Permutation as an enumeration engine. We first identify all $n$ valid pivot nets within the seed netlist. Since each net represents a binary topological choice (swapped or not-swapped), this creates a search space of $2^n$ topologically distinct, functionally equivalent netlists (step \circled{3}). We systematically generate these topologies, capping the number of variants at $\min(100, 2^n)$ per function to maintain a manageable dataset size.

To ensure functional correctness, every generated netlist is rigorously verified. We use \textit{PySpice}\cite{salvaire_pyspice_2021}, an open-source Python library, to simulate each topology and validate its output against the truth table of the original function (step \circled{4}). All enumerated topologies successfully passed this simulation, confirming that LLM-Guided Topology Permutation reliably preserves \textbf{functional equivalence}.

\subsubsection{Dataset Construction}
All enumerated standard cell netlists are processed through a full P\&R flow (step \circled{5}) using \textit{NVCell 2} in 2$nm$ advanced technology nodes. We applied this complete pipeline, from Boolean factoring and topology enumeration to P\&R, to all $254$ non-trivial 3-input functions, yielding this final, topology-diverse corpus totaling $7{,}918$ unique standard cell netlists.

We further curated a specialized dataset composed of all 2,039 \textbf{unroutable} standard cell designs from our corpus, which we then partitioned into training and evaluation sets. By training exclusively on optimizing these unroutable topologies, the LLM learns to identify and penalize the design factors that degrade routability. 

\subsection{GNN Reward Model}\label{subsec:gnn}

Running the full physical design flow to obtain a reward signal is computationally prohibitive in the LLM training process. We therefore use the topology-diverse corpus to train a GNN-based reward model based on \textit{PyTorch Geometric} (PyG)\cite{fey2019fast} (step \circled{6}), which serves as an efficient proxy for P\&R feedback.
A GNN is a natural choice as standard cell netlists are inherently graphs. Routability is fundamentally a topological problem dictated by the pattern of transistor connections. The message-passing mechanism in the GNN is expressly designed to correlate local topological features with global graph properties, making it ideal for predicting routability from the netlist structure of a standard cell.

Our graph construction process is designed to capture the control-flow logic of CMOS netlists. Electrical nets (VDD, GND, I/O, and internal nets) are designated as nodes, with their type (e.g., power or signal) encoded as a node feature. Critically, a transistor is not modeled as a single edge. Instead, we represent its three-terminal nature using two directed edges that capture its specific role and connectivity. The features $\mathbf{e}_{uv}$ on a directed edge between net $u$ and net $v$ encode both the \textbf{transistor type} (e.g., PMOS or NMOS) and the specific \textbf{connection type} (e.g., vdd-to-gate, gate-to-drain, or gate-to-gnd). This fine-grained representation allows the model to accurately identify complex topological features.

We employ a specialized graph convolution layer for the edge-centric structure where message passing is conditional on the edge features. The update rule for a net $v$ at layer $k$ is formulated as:
$$
\mathbf{h}_v^{(k)} = \sigma \left( \frac{1}{|\mathcal{N}(v)|} \sum_{u \in \mathcal{N}(v)} f(\mathbf{e}_{uv}) \cdot \mathbf{h}_u^{(k-1)} \right)
$$
Here, $\mathbf{h}_v^{(k)}$ is the updated feature vector for net $v$, and $\mathbf{h}_u^{(k-1)}$ is the feature vector from neighboring net $u$ at the previous layer. The function $f(\cdot)$ learns a weight matrix conditioned on the edge features $\mathbf{e}_{uv}$, allowing the GNN to process messages differently based on the specific type of transistor connection.
By stacking these layers, the model learns to recognize complex topological patterns, such as long transistor stacks or nets with high diffusion capacitance, that are highly correlated with routing congestion.

We train the model for binary classification with labels $y \in \{-1, 1\}$ (representing unroutable and routable) using a margin-based loss:
$$
\mathcal{L} = \mathbb{E}[\max(0, 1 - y \cdot \hat{y})]
$$
The single logit, $\hat{y}$, is the raw output score of the GNN for the routability classification. This loss function is superior to a standard cross-entropy loss. It pushes the model to produce scores that are not just on the correct side of zero but are far from the decision boundary (e.g., $\hat{y} > 1$ for routable and $\hat{y} < -1$ for unroutable). This creates a large reward gap between good and bad topologies.

The clear and distinct reward signal $\hat{y}$ is highly beneficial for the subsequent group-based advantage calculation in GRPO training. It provides a strong gradient, helping the LLM to efficiently differentiate between routable and unroutable topologies.

\subsection{GRPO Post-Training}\label{subsec:grpo}
\input{algorithm/grpo}

We refine the physical awareness of LLMs for standard cell topology using GRPO, which is particularly effective for this fine-grained optimization for establishing relative baselines through group-level comparisons. This approach is more efficient and stable than traditional methods like PPO by eliminating the need for a separate critic.
The complete post-training process is detailed in Alg.~\ref{alg:grpo_training}.

\subsubsection{Group-Based Advantage Estimation}

The process begins with an instance $\mathbf{q}$, which defines an unroutable standard cell topology(step \circled{7}). Given this instance, the policy $\boldsymbol{\pi}_{\boldsymbol{\theta}_{\text{old}}}$ samples a batch of $M$ candidate nets for swapping, $\mathcal{T} = \{\mathbf{T}_1, \ldots, \mathbf{T}_M\}$ (step \circled{8}). Each of these candidates $\mathbf{T}_j$ is swapped using the LLM-Guided Topology Permutation, resulting in a new netlist $\mathbf{N}_j$ (step \circled{9}). We then evaluate each resulting netlist $\mathbf{N}_j$ using the GNN reward model $\mathcal{R}(\cdot)$ to acquire its routability score, $R_j = \mathcal{R}(\mathbf{N}_j)$ (step \circled{10}). This yields a batch of reward scores, $\mathcal{G} = \{R_1, \ldots, R_M\}$.

To compute the group-based estimated advantage $A_j$, we standardize these rewards within the batch, which effectively contrasts each candidate's performance against its peers (step \circled{11}):
$$
A_j = \frac{R_j - \mu}{\sigma + \epsilon}
$$
where $\mu$ and $\sigma$ are the mean and standard deviation of the rewards in the batch $\mathcal{G}$, and $\epsilon$ is a small constant for stability. This standardization functions as an adaptive baseline, creating a stable, relative gradient signal. It encourages the policy to prioritize actions that generate topologies superior to other options in the batch.

\subsubsection{GRPO Policy Update}

The policy parameters $\boldsymbol{\theta}$ are updated by maximizing the GRPO objective function $\mathcal{L}_{\text{GRPO}}(\boldsymbol{\theta})$. This objective (step \circled{12}) is calculated as a sample mean over $M$ groups, combining a clipped surrogate objective with a KL divergence regularizer:

\[
\begin{aligned}
\mathcal{L}_{\text{GRPO}}(\boldsymbol{\theta})
&=
\frac{1}{M}\sum_{j=1}^M \frac{1}{|\mathbf{T}_j|} \sum_{t=1}^{|\mathbf{T}_j|}
\Biggl\{
\min\!\left\{
\rho_{j,t} A_{j},\;
\operatorname{clip}(\rho_{j,t}, 1-\lambda, 1+\lambda) A_{j}
\right\} \\
&\quad
- \kappa\,\mathcal{D}_{\text{KL}}\!\left(\boldsymbol{\pi}_{\boldsymbol{\theta}} \,\middle\|\, \boldsymbol{\pi}_{\text{ref}}\right)
\Biggr\}
\end{aligned}
\]
Here, the $j$-th candidate net is represented as a sequence of tokens, $\mathbf{T}_j = (o_{j,1}, o_{j,2}, \ldots)$, where $o_{j,t}$ denotes the specific token generated at timestep $t$. The equation $
\rho_{j,t} = \frac{\boldsymbol{\pi}_{\boldsymbol{\theta}}(o_{j,t} \mid q, o_{j,<t})}{\boldsymbol{\pi}_{\boldsymbol{\theta}_{\text{old}}}(o_{j,t} \mid q, o_{j,<t})}
$ is the per-token importance sampling (IS) ratio for $\mathbf{T}_j$.

As shown in Alg.~\ref{alg:grpo_training}, the first term $L_{\text{CLIP}}$ is the clipped surrogate objective, where the hyper-parameter $\lambda$ constrains the probability ratio to prevent excessively large policy updates. The second term $L_{\text{KL}}$ is a soft constraint, scaled by $\kappa$, that regularizes the current policy $\boldsymbol{\pi}_{\boldsymbol{\theta}}$ by penalizing its divergence from a fixed reference policy $\boldsymbol{\pi}_{\text{ref}}$ (a policy pre-trained via behavioral cloning).



%% file: algorithm/dns.tex
\newcommand{\IfThenInline}[2]{%
  \lIf{#1}{#2}%
}
\begin{algorithm}[!t]
\caption{LLM-Guided Topology Permutation}
\label{alg:net_swapping}
\scriptsize
\SetAlgoLined
\SetKwInOut{Input}{Input}
\SetKwInOut{Output}{Output}
\SetKwProg{Fn}{Function}{:}{}%
\SetKwIF{If}{ElseIf}{Else}{if}{:}{else if}{else}{}%
\SetKwFor{ForEach}{foreach}{:}{}%

\SetAlFnt{\normalfont\small}
\newcommand{\mycommentfont}[1]{\small\textrm{#1}}
\SetCommentSty{mycommentfont}

\Input{SPICE netlist of standard cell $L$, pivot net $pivot$ from LLM}
\Output{Modified SPICE netlist $L'$ or Error}

\Fn{SwapNet({L, pivot})}{
    D, G $\leftarrow$ \text{ParseDevicesFrom}(L), \text{BuildGraphFrom}(L)\;
    isPMOSNet, isNMOSNet $\leftarrow false, false$\;
    \ForEach{device $m$ in $D$}{
        \If{($m.\text{drain} =$ pivot \textbf{or} $m.\text{source} =$ pivot)}{
            \IfThenInline{$m.\text{type} =$ \text{PMOS}}{isPMOSNet $\leftarrow true$}
            \IfThenInline{$m.\text{type} =$ \text{NMOS}}{isNMOSNet $\leftarrow true$}
        }
    }
\If{(isPMOSNet \textbf{and} isNMOSNet) \textbf{or} (\textbf{not} isPMOSNet \textbf{and} \textbf{not} isNMOSNet)}
    {
        \Return \text{PromptLLM("Invalid pivot! Please select a new valid net.")}
    }
    UpNeighbor,DownNeighbor $\leftarrow$ G.\text{Neighbors}(pivot)\;
    NCA,NCD $\leftarrow$ \text{BFS}(G, UpNeighbor, DownNeighbor)\;
    UpBoundary $\leftarrow$ G.\text{Neighbors}(NCA) $\cap$ G.\text{Reachable}(pivot, NCA)\;
    DownBoundary $\leftarrow$ G.\text{Neighbors}(NCD) $\cap$ G.\text{Reachable}(pivot, NCD)\;
    $\Delta \leftarrow \emptyset$\;
    \ForEach{device $m$ in $D$}{ 
        $(d, s) \leftarrow (m.\text{drain}, m.\text{source})$\;
        $(d_{new}, s_{new}) \leftarrow (d, s)$\;
        \IfThenInline{$d =$ NCA \textbf{and} $s \in$ UpBoundary}{$d_{new} \leftarrow pivot$}
        \IfThenInline{$d =$ pivot \textbf{and} $s \in$ DownNeighbor}{$d_{new} \leftarrow NCA$}
        \IfThenInline{$s =$ NCD \textbf{and} $d \in$ DownBoundary}{$s_{new} \leftarrow pivot$}
        \IfThenInline{$s =$ pivot \textbf{and} $d \in$ UpNeighbor}{$s_{new} \leftarrow NCD$}
        \IfThenInline{$(d_{new}, s_{new}) \neq (d, s)$}{$\Delta[m] \leftarrow (d_{new}, s_{new})$}
    }
    $L' \leftarrow \text{ApplyChanges}(L, \Delta)$\;
    \Return $L'$\;
}
\end{algorithm}

%% file: algorithm/grpo.tex
\begin{algorithm}[!t]
\caption{GRPO Post-Training in \name}
\label{alg:grpo_training}
\scriptsize
\SetAlgoLined
\SetKwInOut{Input}{Input}
\SetKwInOut{Output}{Output}
\SetKwProg{Fn}{Function}{:}{}%
\SetKwIF{If}{ElseIf}{Else}{if}{:}{else if}{else}{}%
\SetKwFor{ForEach}{foreach}{:}{}%
\SetKwFor{For}{for}{:}{}%

\SetAlFnt{\normalfont\small}
\newcommand{\mycommentfont}[1]{\small\textrm{#1}}
\SetCommentSty{mycommentfont}

\Input{Initial policy $\boldsymbol{\pi}_{\boldsymbol{\theta}}$; reference policy $\boldsymbol{\pi}_{\text{ref}}$; lookup table $\mathcal{R}(\cdot)$; \\ \quad \quad dataset $\mathcal{D}$; group size $M$; training iterations $I$; \\ \quad \quad inner update steps $K$; hyperparameters $\lambda, \kappa, \epsilon$}
\Output{Optimized policy $\boldsymbol{\pi}_{\boldsymbol{\theta}}$}

\BlankLine

\For{$i=1$ \KwTo $I$}{
    $\boldsymbol{\pi}_{\boldsymbol{\theta}_{\text{old}}} \leftarrow \boldsymbol{\pi}_{\boldsymbol{\theta}}$\;
    Sample prompt $\mathbf{q}$ from $\mathcal{D}$ (contains unroutable cell netlist $L$)\;
    $\mathcal{T}, \mathcal{G} \leftarrow \emptyset, \emptyset$\;
    
    \For{$j=1$ \KwTo $M$}{
        $\mathbf{T}_j \sim \boldsymbol{\pi}_{\boldsymbol{\theta}_{\text{old}}}(\cdot \mid \mathbf{q})$\;
        $\mathcal{T} \leftarrow \mathcal{T} \cup \{\mathbf{T}_j\}$\;
    }
    
    \ForEach{$\mathbf{T}_j$ in $\mathcal{T}$}{
        $\mathbf{N}_j \leftarrow$ \text{SwapNet}($L$, $\mathbf{T}_j$)\;
        $R_j \leftarrow \mathcal{R}(\mathbf{N}_j)$\;
        $\mathcal{G} \leftarrow \mathcal{G} \cup \{R_j\}$\;
    }
    
    $\mu, \sigma \leftarrow \text{Mean}(\mathcal{G}), \text{StdDev}(\mathcal{G})$\;
    $\mathcal{A} \leftarrow \emptyset$\;
    \ForEach{$R_j$ in $\mathcal{G}$}{
        $A_j \leftarrow (R_j - \mu) / (\sigma + \epsilon)$\;
        $\mathcal{A} \leftarrow \mathcal{A} \cup \{A_j\}$\;
    }

    \For{$k=1$ \KwTo $K$}{
        $L_{\text{CLIP}} \leftarrow \min\!\left\{
            \rho_{j,t} A_j,\;
            \operatorname{clip}(\rho_{j,t}, 1-\lambda, 1+\lambda) A_j
            \right\}$\;
        $L_{\text{KL}} \leftarrow \mathcal{D}_{\text{KL}}\!\left(\boldsymbol{\pi}_{\boldsymbol{\theta}} \,\middle\|\, \boldsymbol{\pi}_{\text{ref}}\right)$\;
        $\mathcal{L}_{\text{GRPO}}(\boldsymbol{\theta}) \leftarrow \frac{1}{M}\sum_{j=1}^M \frac{1}{|\mathbf{T}_j|} \sum_{t=1}^{|\mathbf{T}_j|} (L_{\text{CLIP}} - \kappa\ L_{\text{KL}})$;\
        
        Update $\boldsymbol{\theta}$ by maximizing $\mathcal{L}_{\text{GRPO}}(\boldsymbol{\theta})$\;
    }
}
\Return $\boldsymbol{\pi}_{\boldsymbol{\theta}}$\;

\end{algorithm}

%% file: 05sec-results.tex
\newcommand{\impr}[1]{~{\color{red}+#1}}
\newcommand{\decr}[1]{~{\color{red}-#1}}
\begin{table}[htbp]
\centering
\caption{Comparative analysis of model performance in the evaluation benchmark. The overall SOTA results are shown in \textbf{bold}. Numbers in {\color{red}red} indicate the performance improvement of our fine-tuned \name models over the respective Qwen2.5-Coder base models.}
\label{tab:my-model-comparison}
\resizebox{0.95\linewidth}{!}{%
\begin{tabular}{lll}
\toprule
\textbf{Name} & \textbf{PDA Congestion (Mean)} & \textbf{Routable Rate (\%)} \\
\midrule
Baseline: SO3-Cell & N/A & N/A \\
Ours: \name-3B & 4.06\decr{0.26(6.02\%)} & 69.7\impr{14.1(25.4\%)} \\
Ours: \textbf{\name-7B} & \textbf{3.90}\decr{0.34(8.02\%)} & \textbf{77.3}\impr{24.3(45.8\%)} \\
\midrule
Qwen2.5-Coder-3B & 4.32 & 55.6 \\
Qwen2.5-Coder-7B & 4.24 & 53.0 \\
Qwen2.5-Coder-32B & 4.33 & 54.0 \\
\midrule 
Qwen3-4B & 4.19 & 50.5 \\
Qwen3-8B & 4.34 & 59.1 \\
Qwen3-32B & 4.11 & 58.1 \\
\midrule 
CodeLlama-7b & 4.32 & 48.0 \\
CodeLlama-34B & 4.23 & 49.5 \\
Llama-3.3-70B & 4.31 & 56.1 \\
\midrule 
DeepSeek-Coder-33B & 4.10 & 58.6 \\
DeepSeek-R1-0528 & 4.12 & 51.5 \\
DeepSeek-V3.2-Exp & 4.26 & 56.6 \\
\midrule 
Codestral-22B & 4.26 & 55.1 \\
Mistral Medium 3.1 & 4.29 & 58.1 \\
\midrule 
GPT-5-Codex & 4.13 & 54.5 \\
GPT-5 & 4.19 & 56.1 \\
\midrule 
Gemini 2.5 Pro & 4.21 & 54.6 \\
\midrule 
Claude Opus 4.1 & 4.25 & 55.1 \\
\bottomrule
\end{tabular}%
}
\end{table}

\begin{table*}[t!]
\centering
\small 
\captionsetup{skip=10pt} 
\caption{Comparison of layout quality and runtime between SO3-Cell and \name.
\name generates optimized netlists in only a fraction of the time required by SO3-Cell, achieving comparable layout quality with an average speedup of 85.91$\times$.}
\label{tab: compSO3}
\begin{tabular}{l r r r r r r r r r}
\toprule
\multirow{2}{*}{\textbf{Cell}} & {\multirow{2}{*}{\textbf{\#$I$}}} & {\multirow{2}{*}{\textbf{\#$T$}}} & \multicolumn{2}{c}{\textbf{SO3-Cell}} & \multicolumn{5}{c}{\textbf{\name}} \\ 
\cmidrule(lr){4-5} \cmidrule(lr){6-10}
 & & & {\textbf{OTC$\downarrow$}} & {\textbf{Runtime (sec)}} & {\textbf{OTC$\downarrow$}} & {\textbf{$t_{LLM}$ (sec)}} & {\textbf{$t_{P\&R}$ (sec)}} & {\textbf{Runtime (sec)}} & {\textbf{Speedup ($\times$)}} \\
\midrule
NAND4\_X2 & 4 & 8 & \textbf{685} & 3258.875 & \textbf{685} & 1.843 & 187.744 & 189.587 & \textbf{17.19} \\
NOR4\_X2 & 4 & 8 & \textbf{685} & 10176.047 & \textbf{685} & 1.793 & 149.688 & 151.481 & \textbf{67.18} \\
AOI211\_X1 & 4 & 8 & \textbf{53} & 24.683 & \textbf{53} & 1.803 & 0.654 & 2.457 & \textbf{10.05} \\
OAI211\_X1 & 4 & 8 & \textbf{53} & 60.099 & 85 & 1.762 & 14.077 & 15.839 & \textbf{3.79} \\
AOI22\_X2 & 4 & 8 & \textbf{355} & 14733.220 & \textbf{355} & 1.710 & 1139.043 & 1140.753 & \textbf{12.92} \\
OAI22\_X2 & 4 & 8 & \textbf{355} & 24431.104 & \textbf{355} & 1.714 & 1469.195 & 1470.909 & \textbf{16.61} \\
AOI221\_X1 & 5 & 10 & \textbf{149} & 258.367 & \textbf{149} & 1.882 & 5.438 & 7.320 & \textbf{35.30} \\
OAI221\_X1 & 5 & 10 & \textbf{149} & 126.102 & \textbf{149} & 1.932 & 9.132 & 11.064 & \textbf{11.40} \\
AOI222\_X1\_SH & 6 & 12 & \textbf{124} & 35488.258 & \textbf{124} & 1.796 & 61.347 & 63.143 & \textbf{562.03} \\
OAI222\_X1\_SH & 6 & 12 & \textbf{124} & 8066.431 & \textbf{124} & 1.902 & 63.867 & 65.769 & \textbf{122.65} \\
\bottomrule
\end{tabular}
\end{table*}

\section{EXPERIMENTS}\label{sec:results}

\subsection{Experiment Setup}

We implement the GRPO algorithm within \name, leveraging the \textit{Verl} library~\cite{sheng2025hybridflow} and the \textit{SGLang} serving framework~\cite{zheng2024sglang}. All experiments are conducted on an NVIDIA DGX Station equipped with four 80GB A100 GPUs, using Qwen2.5-Coder-3B and Qwen2.5-Coder-7B as base models. Both models are trained for 15 epochs with a batch size of 256 and a default GRPO learning rate of $10^{-6}$. In this configuration, a single training epoch takes approximately 30 minutes for the 3B model and 70 minutes for the 7B model.

\vspace{-1mm}

\subsection{Routability Optimization}\label{subsec:benchmark-results}
Table~\ref{tab:my-model-comparison} benchmarks \name against several other foundation models~\cite{hui2024qwen2, yang2025qwen3, roziere2023code, grattafiori2024llama, guo2025deepseek, guo2024deepseek, liu2024deepseek, comanici2025gemini, achiam2023gpt, jiang2023mistral} using routing results from \textit{NVCell 2} on an advanced 2$nm$ technology node.
The foundation models performed this task without any fine-tuning. 

The evaluation employs two key metrics: \textbf{PDA Congestion} and \textbf{Routable Rate}. \textit{PDA Congestion}, reported as the average value across the benchmark, is a novel DTCO metric introduced in \textit{NVCell 2} that quantifies local congestion hotspots. It models the density of crossing nets and the required contacts to the lowest metal layer. In standard cell layout, lower \textit{PDA Congestion} indicates a cell topology with fewer local congestion issues, which directly mitigates routing failures and improves routability. While \textit{Routable Rate} measures the percentage of standard cells, originally unroutable, that \textit{NVCell 2} successfully routes following LLM-Guided Topology Permutation. 

The results highlight the efficacy of our GRPO post-training. Our \textbf{\name-7B} model achieves a \textbf{77.3\%} \textit{Routable Rate} and a \textbf{3.90} mean \textit{PDA Congestion}. This marks a substantial 24.3 percentage point (45.8\%) improvement in routability and an 8.02\% reduction in congestion over the base model. The smaller \textbf{\name-3B} model exhibits similarly strong gains, boosting routability by 14.1 percentage points (25.4\%) and cutting congestion by 6.02\% compared to its own base. These findings validate that the GRPO process successfully learns to optimize standard cell topology by incorporating physical awareness. This optimization mitigates pin-density congestion, leading to a significantly higher \textit{Routable Rate}.

Notably, our specialized \textbf{\name-3B} (\textbf{69.7\%} routable) drastically outperforms much larger models like DeepSeek-V3.2-Exp (56.6\%) and GPT-5 (56.1\%). These general-purpose models, while powerful, produce netlists with high \textit{PDA congestion} (4.26 and 4.19, respectively). This confirms that general-purpose reasoning is insufficient. Domain-specific, physical-aware optimization via GRPO is the key factor in transforming standard cell topology.

In contrast, \textit{SO3-Cell}, the SOTA standard cell generation framework, suffers from a clear scalability bottleneck. Its exhaustive search for topology optimization proved computationally intractable for the evaluation benchmark, which consists entirely of designs with 16 or more transistors. As a result, it failed to generate an optimal topology for any standard cell within the one-hour timeout.

\vspace{-1mm}

\subsection{Integrated Application to SO3-Cell}
To further evaluate the effectiveness of \name, we integrated it into \textit{SO3-Cell}. \textit{SO3-Cell} jointly optimizes cell topology, placement, and routing, but its core recursive design-space exploration for topology optimization incurs a prohibitive computational cost.
We replaced this computationally expensive exploration module with \name to perform topology optimization. 

Table~\ref{tab: compSO3} compares our \textbf{\name-7B} model (see Section~\ref{subsec:benchmark-results}) with baseline \textit{SO3-Cell}. Both methods generated 7$nm$ standard cells with $\geq$4 input signals under an identical setup. We evaluate performance using layout quality, measured by the \textbf{Optimal Total Cost} (\textit{OTC}), and end-to-end \textbf{Runtime}. \textit{OTC} is a comprehensive DTCO objective from \textit{SO3-Cell}. For our method, \textit{Runtime} is defined as $t_{LLM}+t_{P\&R}$, combining topology inference ($t_{LLM}$) and physical layout ($t_{P\&R}$) time. The ``\#$I$'' and ``\#$T$'' columns list the number of input signals and the transistor count, respectively, for the corresponding standard cell. The results show that \name achieves layout quality on par with \textit{SO3-Cell} (comparable \textit{OTC} in all cases except OAI211\_X1) while delivering an average end-to-end speedup of \textbf{85.91$\times$}. This highlights the significant efficiency gains from \name.

Notably, \name demonstrates strong zero-shot generalization. It was trained exclusively on 3-input Boolean functions using a 2$nm$ technology, with no exposure to the target 7$nm$ cell libraries. Despite this, it produces high-quality layouts for these unseen cells in 7$nm$ library and effectively scales to more complex 4-, 5-, and 6-input functions, achieving quality comparable to the baseline. This capability underscores that \name successfully learns transferable design principles without further task-specific fine-tuning.


\subsection{GRPO vs. SFT}

\begin{figure}[h!]
    \centering
    \includegraphics[width=0.9\linewidth]{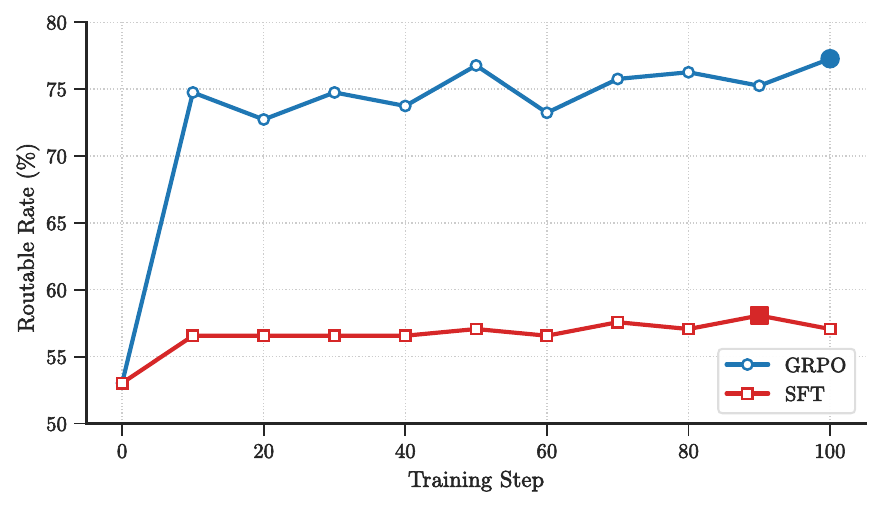}
    \caption{Comparison of Routable Rate in the evaluation benchmark during training using GRPO and SFT. Solid markers denote peak performance for each method.}
    \label{fig: training}
    \vspace{-1mm}
\end{figure}


 We benchmark our GRPO-based approach against a SFT baseline to further evaluate its efficacy. We establish a fair comparison by ensuring both methods use the same Qwen2.5-Coder-7B base model, the same training dataset, and identical training parameters. For the SFT training, the expected ``golden answer'' is set as the first net swap that yields the highest reward according to the GNN model.

Fig.~\ref{fig: training} clearly illustrates the fundamental limitations of SFT for this task. The model's training curve using SFT quickly saturates, hitting a performance ceiling with a \textit{Routable Rate} of only \textbf{57\%} on the evaluation benchmark. This plateau indicates that SFT is limited to imitating its training data. It cannot explore the design space to discover novel, superior solutions. In stark contrast, the GRPO policy demonstrates the power of active exploration and refinement. Its \textit{Routable Rate} rapidly jumps to \textbf{75\%} within just 10 optimization steps and continues to climb, ultimately peaking at over \textbf{77\%}. This rapid ascent shows that GRPO is effectively optimizing standard cell topologies by navigating the design space, far surpassing the performance ceiling defined by the SFT dataset.


This result demonstrates that standard cell topology optimization is particularly well-suited for GRPO post-training. The optimization landscape is such that multiple topological modifications can yield comparable routability improvements. There is no singular ``golden answer'' required by SFT.
The robust nature of GRPO, combined with its ability to optimize for a reward signal in the absence of a single supervised target, makes it the ideal choice for this task.

%% file: 06-sec-conclusion.tex
\section{Conclusion}\label{sec:conclusion}

In this paper, we introduce \name, a novel LLM framework that overcomes scalability bottlenecks in standard cell topology optimization. By leveraging GRPO post-training , we align LLMs with P\&R feedback to efficiently discover physically-aware, routable topologies. \name demonstrates excellent performance while exhibiting robust zero-shot generalization, scaling from 2nm training to more complex 7nm cells. Our evaluations confirm \name achieves comparable layout quality to the SOTA framework along with a significant speedup. This work establishes a new direction for AI-driven design automation, enabling deeper DTCO at the standard cell level.

\textbf{Acknowledgment} -- This work is supported by National Science Foundation (NSF) under CCF2349670, CCF2428520, CCF2403134 awards.